\documentclass[runningheads]{llncs}
\usepackage[T1]{fontenc}
\usepackage[misc]{ifsym}
\usepackage[table]{xcolor}
\usepackage{amsmath}
\usepackage{amssymb}
\definecolor{headercolor}{gray}{0.85}
\usepackage[colorlinks=true,linkcolor=black,citecolor=blue,urlcolor=blue,]{hyperref}
\usepackage{graphicx}
\usepackage{multirow}

\begin{document}
\title{SAM-Med3D-MoE: Towards a Non-Forgetting Segment Anything Model via Mixture of Experts for 3D Medical Image Segmentation}
\titlerunning{SAM-Med3D-MoE}
\newcommand*\samethanks[1][\value{footnote}]{\footnotemark[#1]}
\author{Guoan Wang\inst{1,2}\thanks{Equal contribution,  $^\dagger$ Corresponding author.} \and
Jin Ye\inst{1,3}\samethanks \and
Junlong Cheng\inst{1,4} \and
Tianbin Li \inst{1}\and
Zhaolin Chen \inst{3}\and
Jianfei Cai  \inst{3}\and
Junjun He \inst{1}\textsuperscript{$\dagger$} \and
Bohan Zhuang \inst{3}\textsuperscript{$\dagger$}}
%
\authorrunning{G. Wang et al.}
\institute{Shanghai Artificial Intelligence Laboratory, Shanghai, China \and
College of Computer Science, East China Normal University, Shanghai, China\and
Department of Data Science and AI, Faculty of IT, Monash University, Australia \and
College of Computer Science, Sichuan University, Chendu, China
}
\maketitle            
\begin{abstract}
Volumetric medical image segmentation is pivotal in enhancing disease diagnosis, treatment planning, and advancing medical research. While existing volumetric foundation models for medical image segmentation, such as SAM-Med3D and SegVol, have shown remarkable performance on general organs and tumors, their ability to segment certain categories in clinical downstream tasks remains limited. Supervised Finetuning (SFT) serves as an effective way to adapt such foundation models for task-specific downstream tasks but at the cost of degrading the general knowledge previously stored in the original foundation model.To address this, we propose SAM-Med3D-MoE, a novel framework that seamlessly integrates task-specific finetuned models with the foundational model, creating a unified model at minimal additional training expense for an extra gating network. This gating network, in conjunction with a selection strategy, allows the unified model to achieve comparable performance of the original models in their respective tasks — both general and specialized — without updating any parameters of them.Our comprehensive experiments demonstrate the efficacy of SAM-Med3D-MoE, with an average Dice performance increase from 53.2\% to 56.4\% on 15 specific classes. It especially gets remarkable gains of 29.6\%, 8.5\%, 11.2\% on the spinal cord, esophagus, and right hip, respectively. Additionally, it achieves 48.9\% Dice on the challenging SPPIN2023 Challenge, significantly surpassing the general expert's performance of 32.3\%. We anticipate that SAM-Med3D-MoE can serve as 
a new framework for adapting the foundation model to specific areas in medical image analysis. Codes and datasets will be publicly available.


\end{abstract}
\begin{figure}\centering
\includegraphics[width=\textwidth]{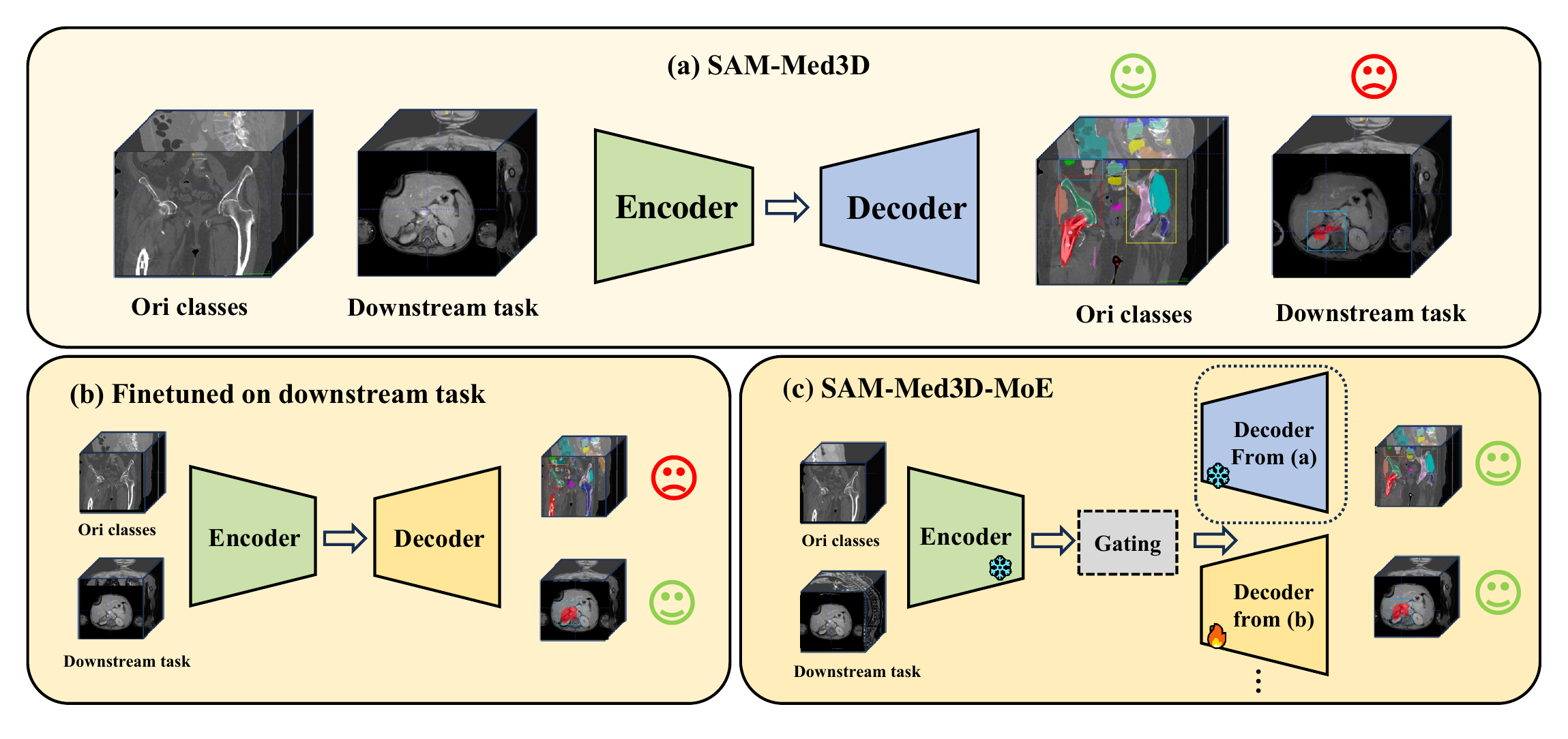}
\caption{Advantages of SAM-Med3D-MoE in general tasks and specific downstream tasks. 
(a) SAM-Med3D, a foundational model for volumetric medical image segmentation, demonstrates remarkable performance in segmenting general organs and tumors. However, its performance is notably less effective in segmenting neuroblastoma as observed in the SPPIN2023 challenge.
(b) After finetuning SAM-Med3D on the SPPIN2023, it enhanced its performance on neuroblastoma segmentation but diminished its overall segmentation capability.
(c) Our method is competent for both general and downstream tasks.   
} \label{fig1}
\end{figure}

\section{Introduction}
Volumetric medical image segmentation is a fundamental task in 3D medical image analysis, which plays a crucial role in diagnosing, radiotherapy planning, treating, and further medical research~\cite{antonelli2022medical,NEURIPS2022_ee604e1b,Ma-2021-AbdomenCT-1K}. 
Compared to the traditional manual segmentation by specialists, deep learning-based 3D medical image segmentation models~\cite{huang2023stunet,isensee2021nnu,ronneberger2015u} can achieve accurate results in several clinical scenarios. However, these models are designed and trained on task-specific data, leading to a significant decline in performance when applied to new tasks or different imaging modalities.

With the vast computational resources available and large amounts of labeled data, 
the demand for universal foundation models in medical image segmentation is intensely growing~\cite{ma2024segment}. Such models can be trained once and then applied to a wide range of segmentation tasks.
Recently, Segment Anything Model (SAM)~\cite{kirillov2023segment}, a promptable foundation model in natural image segmentation, has overcome the limitations of traditional specialist models that rely on fully supervised learning on task-specific data and demonstrated remarkable performance in zero-shot scenarios.  Due to the great success of SAM, attempts have been made~\cite{du2024segvol,wang2023sammed3d} to build foundation models for 3D medical image segmentation, e.g., SAM-Med3D~\cite{wang2023sammed3d}, via training across a vast collection of public datasets (over 100k volumetric masks). 

Although these foundation models have achieved noticeable performance gains on most publicly accessible data pertinent to organs and tumors, they are still difficult to directly apply to practical deployments. 
As shown in Fig.~\ref{fig1} (a), while SAM-Med3D~\cite{wang2023sammed3d} can perform general medical image segmentation, it still struggles with new specific tasks (e.g., to segment neuroblastoma in MRI data).The inherent reason stems from the lack of large-scale publicly accessible data due to the unique challenges of privacy and strictly ethical issues in medical imaging. 
Even though SegVol~\cite{du2024segvol} and SAM-Med3D~\cite{wang2023sammed3d} have consolidated hundreds of publicly accessible datasets, 
resulting in 5.7k images with 149k corresponding masks and another 21k images with 131k corresponding masks, these numbers amount to merely about 0.1~\% of images and 0.01~\% of masks used in training SAM. Moreover, the diversity of the existing public datasets for medical images is limited, rendering such models difficult to address clinical downstream tasks that fall outside the scope of the datasets. For example, each year's MICCAI Challenge introduces new segmentation demands within the field of medical image segmentation, such as SPPIN2023~\cite{buser2023surgical}, which focuses on the new task of segmenting neuroblastoma in children's MRI scans.

Supervised Finetuning (SFT) is crucial for efficiently adapting foundation models for task-specific downstream tasks~\cite{cheng2023sammed2d,chung2022scaling,hu2021lora}. While finetuning foundation models with task-specific data can enhance their performance on downstream tasks, it would inadvertently degrade the general knowledge previously stored in foundation models~\cite{dou2024loramoe} as shown in Fig.~\ref{fig1} (b). Thus, in this paper, our motivation is to devise a method that can seamlessly integrate the original foundation model with task-specific finetuned models into a supernet, which is proficient in both general and specific tasks.

Recently, MoE (Mixture of Experts) ~\cite{jacobs1991adaptive,lepikhin2020gshard,shazeer2017outrageously} has become popular in assembling several expert models into one powerful foundation model for LLMs~\cite{fedus2022switch,jiang2024mixtral}. 
Inspired by MoE, we propose the Segment Anything Model on 3D Medical images with Mixture of Experts (SAM-Med3D-MoE), which assembles any task-specific finetuned model (specific expert) with the foundational model (general expert) to a new model, at a cheap cost of training an extra lightweight gating network as shown in Fig.~\ref{fig1} (c). 
Specifically, our approach utilizes a gating network that processes both image and prompt embeddings to generate confidence scores for each specific expert. We further introduce a novel selection strategy that adaptively combines the outputs from the general expert and the Top-1 specific expert to yield the final mask.

In summary, the contributions of this paper can be summarized as follows. 
(1) SAM-Med3D-MoE is the first to introduce MoE techniques to adaptively merge the general knowledge from the foundational model and specific domain knowledge from task-specific finetuned models for volumetric medical image segmentation.(2) We introduce a lightweight, trainable gating network and a selector module designed to expand foundation models for downstream tasks. (3) We evaluate the effectiveness of SAM-Med3D-MoE on the SPPIN MICCAI 2023 Challenge and 15 existing classes where the foundation model was inferior to specific expert models. 
The extensive experiments demonstrate the efficacy of SAM-Med3D-MoE, with an average Dice performance increase from 53.2\% to 56.4\% on 15 specific classes, it especially gets remarkable gains of 29.6\%, 8.5\%, 11.2\% on the spinal cord, esophagus, right hip, respectively. Additionally, it achieves 48.9\% Dice on the challenging SPPIN2023 Challenge, significantly surpassing the general expert's performance of 32.3\%. 

\begin{figure}\centering
\includegraphics[width=\textwidth]{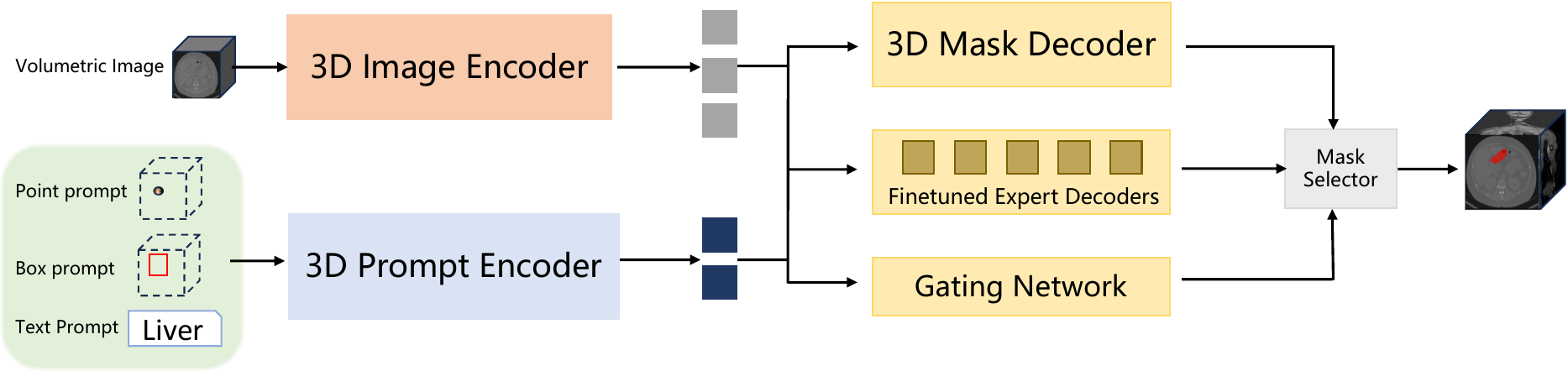}
\caption{Overview of our SAM-Med3D-MoE approach. The outputs of 3D Image Encoder and Prompt Encoder undergo the dynamic selection by the gating mechanism. If the weight of the top-1 selection after softmax exceeds $\tau$, the most proficient finetuned expert decoder (specific expert) is chosen, together with 3D Mask Decoder (general expert). Conversely, if the weight does not exceed $\tau$, only 3D Mask Decoder is utilized.} \label{fig2}
\end{figure}

\section{Method}
Our model is built upon SAM-Med3D~\cite{wang2023sammed3d}, which can be decoupled into three parts: \textbf{1) 3D Image Encoder} that is based on ViT (Vision Transformer)~\cite{vit},  
a much stronger backbone than convolutional encoders when trained on large-scale datasets; \textbf{2) Prompt Encoder} to handle both point and box prompts, which are represented using frozen 3D absolute positional encodings and then combined with learned embeddings specific to each prompt type; \textbf{3) 3D Mask Decoder}, a lightweight module to efficiently map the image embedding and prompt embeddings to an output mask. In the following sections, we present the details of our proposed SAM-Med3D-MoE.

\subsection{Overview of SAM-Med3D-MoE}
The unified framework is composed of a general expert alongside several task-specific experts, the latter being finetuned on the 3D mask decoder alone. This setup enables the use of the identical 3D image encoder and 3D prompt encoder throughout the model. For the 3D mask decoders, we distinguish them into two categories: the general expert (i.e., 3D Mask Decoder in Fig.~\ref{fig2}) and the task-specific experts (i.e., Finetune Expert Decoders in Fig.~\ref{fig2}). 
Then, a gating network is adopted to process both image and prompt embeddings to generate confidence scores for each task-specific expert, and we further introduce a novel selection strategy that adaptively combines the outputs from the general expert and the Top-1 specific expert to yield the final mask.

\subsection{Gating Network}
\begin{figure}\centering
\includegraphics[width=0.9\textwidth]{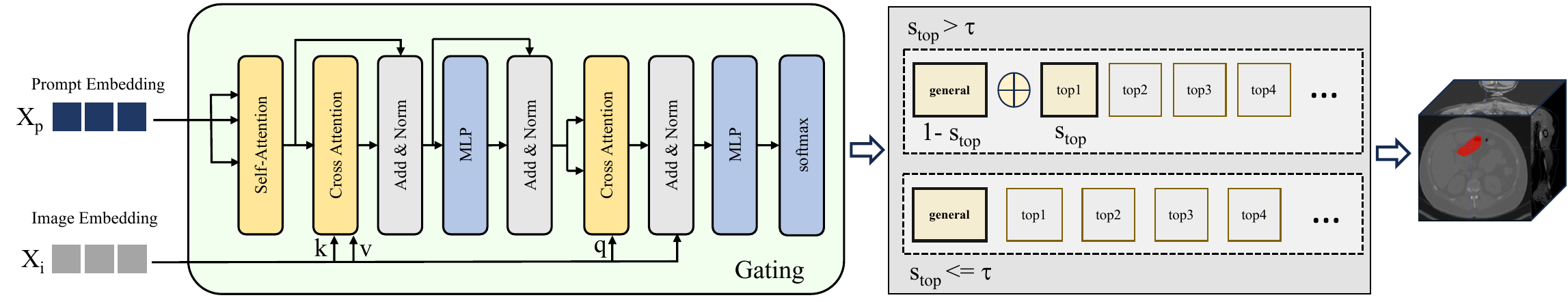}
\caption{Details of the gating network and the selector.
} \label{fig3}
\end{figure}

As shown in Fig.~\ref{fig3}, the gating network is responsible for calculating the confidence score for every expert model. Specifically, we take image embedding $X_{i}\in\mathbb{R}^{\frac{HWD}{16^3} \times C}$ and prompt embedding $X_{p}\in\mathbb{R}^{C}$ as input. First, the prompt embedding goes through a self-attention and output as a query to engage in cross-attention with $X_{i}$ (as the key and value), thereby establishing a correlation between the prompt and the image. Then, an MLP layer is adopted to update the prompt embedding, and its result is used as the key and value to inject its information into image embedding (as the query) with cross-attention. Notably, residual connections and normalization layers are added after each attention and MLP layer. Last, we send the output feature to two successive fully connected layers and a softmax layer to obtain final scores $S\in\mathbb{R}^{m}$ for $m$ experts. 

\subsection{Mask Selector}
\label{Mask Selector}
Following the gating network, we assign a confident weight to each expert's output. To significantly prevent the model from forgetting its original segmentation capabilities, we introduce a hyper-parameter $\tau$ that acts as a switch, which allows the model to select expert models only if the top score exceeds a predetermined threshold. When the switch is activated, rather than exclusively choosing the output of the first-ranked expert $M_{top}\in\mathbb{R}^{H \times W \times D}$, we simply implement weighted sum to fuse it with the general output $M_{g}\in\mathbb{R}^{H \times W \times D}$. The whole progress can be formulated as below: 
\begin{align*}
    M_o = 
    \begin{cases}
        (1 - s_{top}) \times M_g + s_{top} \times M_{top} & \quad s_{top} > {\tau} \\
        M_g & \quad s_{top} \le {\tau} \\
    \end{cases}
\end{align*}
where $M_o\in\mathbb{R}^{H \times W \times D}$ is the final mask and $s_{top}$ is the confidence score of the first-ranked expert.

\section{Experiments and Discussion}

\noindent\textbf{Implementation Details.} The SAM-Med3D-MoE architecture underwent training within the PyTorch deep learning framework, exhibiting memory usage that scaled with the count of experts incorporated. Specifically, for the variant comprising 15 MoE experts, we employ 8 Nvidia V100 GPUs, each furnished with 32 GB of RAM. 
Despite the increasing memory requirements due to the multiplicity of experts, it is worth mentioning that the training speed remains nearly on par with that of the baseline SAM-Med3D model. This efficiency is attributable to our strategic training strategy, wherein we freeze the parameters of both the image and prompt encoders, confining updates exclusively to the parameters of the Top-1 expert mask decoder.


Our chosen loss function, DiceCELoss, is applied on top of the final predictive results, while CrossEntropy Loss is utilized to supervise the outputs of the gating mechanism, thereby ascertaining the accurate selection of the appropriate expert. We set the learning rate to $1 \times 10^{-4}$ for fine-tuning the experts and $1 \times 10^{-6}$ for the training of the gating network. The AdamW optimizer is employed for the optimization of parameters. For fair comparisons, the dataset employed for training is the same as that used for the initial baseline SAM-Med3D model.

\subsection{Experiments}
\noindent\textbf{Extensions on Downstream Tasks.}
To extend the model to downstream tasks, traditional methods typically finetune the pretrained model partially or entirely on the new task. However, this may lead to the model ``forgetting'' the knowledge acquired on the original task, a phenomenon we refer to as ``catastrophic forgetting''. Our SAM-Med3D-MoE can effectively alleviate this problem. Specifically, we conduct our experiments on the SPPIN MICCAI 2023 Challenge~\cite{buser2023surgical}, which is a dataset that SAM-Med3D has never encountered during the training process.
As shown in Table~\ref{tab1}, the task-specific finetuned expert significantly improved the performance of the baseline model (by approximately 17\%). However, this finetuned model encountered difficulties in adapting to the original task (as shown in the third row of the left half of Table~\ref{tab1}, ``Ori tasks'' refers to the original tasks that the Baseline SAM-Med3D had previously learned), resulting in its performance being lower than the baseline.
Our SAM-Med3D-MoE address this issue by adding an expert on top of the baseline network to adapt to the SPPIN dataset. The advantage of this approach lies in the fact that by only training the gating network, we can achieve performance improvements on the new SPPIN task while maintaining stable or slightly decreased performance on the original task. This demonstrates the effectiveness of our SAM-Med3D-MoE in mitigating catastrophic forgetting and enabling the model to adapt to new tasks without compromising its performance on the original task.

\begin{table}[ht]\centering
\caption{The comparison of the Dice scores on downstream tasks and weak categories, including the prompt as 6 points (on the left) and bbox (on the right). Our method mitigates catastrophic forgetting and enables the model to adapt to new tasks, while having minimal impact on the performance of the original task. The \textbf{bold} content is the highest value.}\label{tab1}
\setlength{\tabcolsep}{1.3mm}{
\begin{tabular}{l|clcl|llcl}
\hline
\multirow{2}{*}{Model} & \multicolumn{4}{c|}{Downstream Task (Point/Bbox)}                   & \multicolumn{4}{c}{Weak Categories (Point/Bbox)}                              \\ \cline{2-9} 
                       & \multicolumn{2}{c|}{Ori tasks} & \multicolumn{2}{c|}{SPPIN}       & \multicolumn{2}{c|}{Other classes} & \multicolumn{2}{c}{finetune 15 classes} \\ \hline
Baseline               & \multicolumn{2}{c|}{ \textbf{0.433}/\textbf{0.527}} & \multicolumn{2}{c|}{0.338/0.323} & \multicolumn{2}{l|}{\textbf{0.424}/\textbf{0.541}}   & \multicolumn{2}{c}{0.399/0.532}          \\
FT-expert              & \multicolumn{2}{c|}{0.333/0.438} & \multicolumn{2}{c|}{\textbf{0.503}/\textbf{0.510}} & \multicolumn{2}{l|}{0.036/0.094}   & \multicolumn{2}{c}{\textbf{0.660}/\textbf{0.637}}          \\ \hline
Ours                   & \multicolumn{2}{c|}{\textcolor{red}{0.411}/\textcolor{red}{0.527}} & \multicolumn{2}{c|}{\textcolor{red}{0.451}/\textcolor{red}{0.489}} & \multicolumn{2}{l|}{\textcolor{red}{0.353}/\textcolor{red}{0.400}}   & \multicolumn{2}{c}{\textcolor{red}{0.520}/\textcolor{red}{0.564}}          \\ \hline
\end{tabular}
}
\end{table}

\begin{figure}\centering
\includegraphics[width=\textwidth]{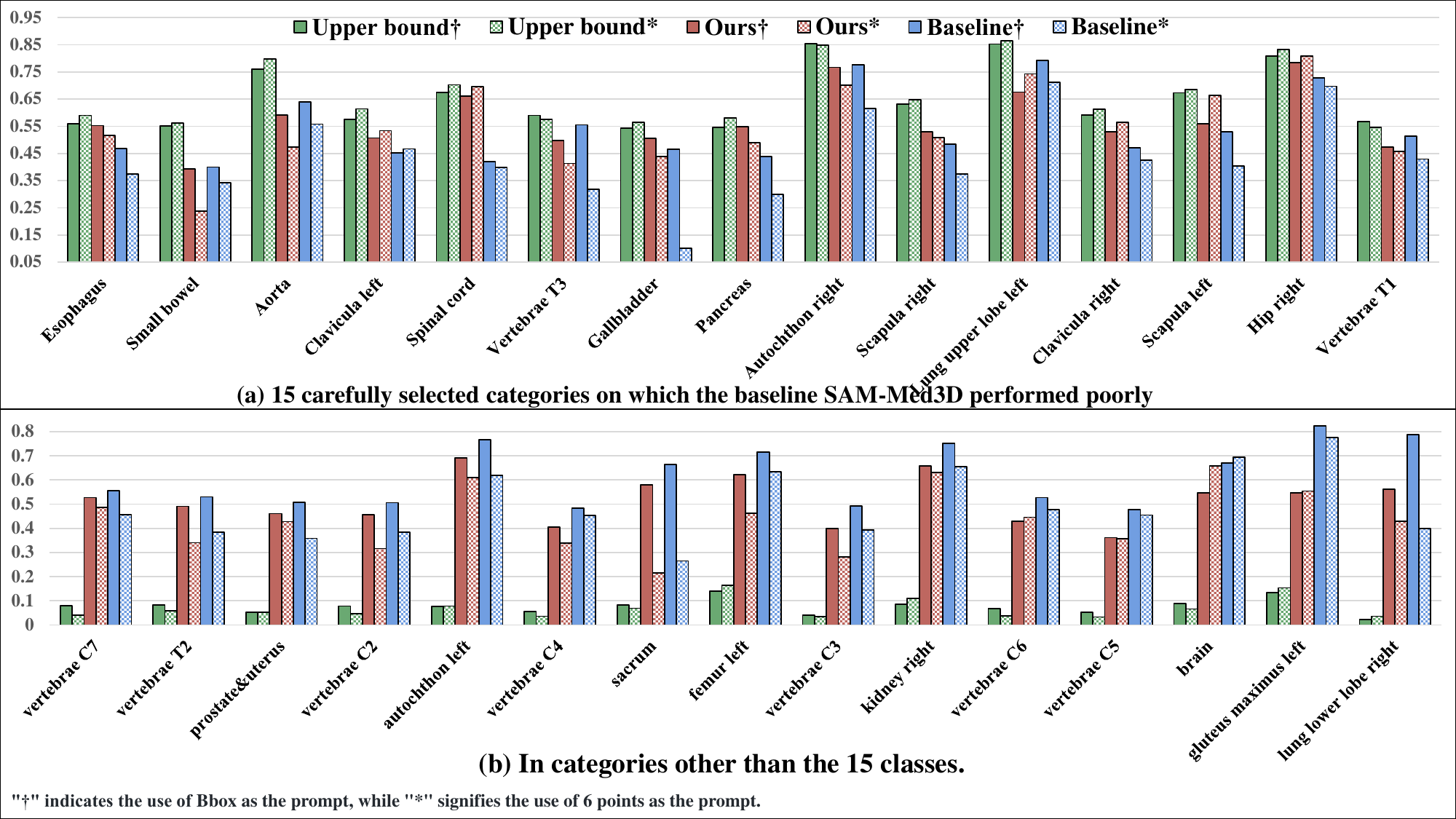}
\caption{Our model exhibits strong performance on (a) the 15 selected categories as well as on (b) the original un-finetuned categories.
} \label{fig4}
\end{figure}

\noindent\textbf{Extensions on Weak Categories.} Fig.~\ref{fig4} illustrates the comparative accuracies of the baseline SAM-Med3D, the finetuned model (denoted as the Upper bound), and our proposed SAM-Med3D-MoE. Panel (a) delineates 15 categories meticulously chosen based on the subpar performance of the baseline SAM-Med3D. To enhance performance on these categories, we dedicated an expert model to each, resulting in substantially improved accuracies, as (a) attests.Despite these gains, a notable drawback emerges, as (b) reveals: models finetuned on isolated categories tend to overfit, losing generalizability and thus underperforming across the broader category spectrum. In response, we amalgamated the expert models for the 15 categories within a MoE framework. Subsequent fine-tuning of the MoE's gating network yielded a model whose segmentation prowess notably eclipsed that of the individually finetuned counterparts.
Crucially, as (b) corroborates, the integration into the SAM-Med3D-MoE did not detrimentally impact performance on the baseline categories. This outcome underscores the efficacy of the gating network in judiciously selecting the relevant expert, circumventing the pitfalls intrinsic to conventional fine-tuning approaches. For a comprehensive assessment, the collective average test scores are summarized in Table~\ref{tab1}. The phrase ``finetune 15 classes'' pertains to the specifically chosen categories upon which we conducted fine-tuning. Conversely, ``Other classes'' represent the residual categories within the validation dataset that were not included in the selected 15 for fine-tuning. \par

\subsection{Ablation Study}
To ascertain the most efficacious configuration of the mask selector, we undertook evaluations across four anatomical categories: esophagus, small bowel, stomach, and aorta. Detailed in Table~\ref{tab2}, our examination spanned six distinct scenarios: the baseline SAM-Med3D, four category-specific finetuned models representing the upper bound, and two variants employing thresholds ($\tau$) of 0.5 and 0.7. In these latter scenarios, we substituted the weighted sum with an arithmetic mean (avg) and refined the weighted sum formula, transitioning from $s_{top}$ to a softmax-fused output of the top expert mask decoder and the general model's decoder($Aft_{weight}$). For both variants, we maintained a constant $\tau$ of 0.5. Our findings revealed that the gating mechanism's proficiency in assimilating cues from the input image and prompt information significantly bolsters the model's accuracy. This enhancement is particularly evident when the mask selector capitalizes on $s_{top}$ feature information within the weighted sum approach. Pertaining to the threshold $\tau$, we discerned that elevating $\tau$ diminishes accuracy, as a higher $\tau$ may inadvertently bias the model towards the general decoder, thereby compromising precision.\par

\begin{table}[ht]\centering
\caption{The comparison of the Dice scores for evaluating different configurations in mask selector across four categories, including the prompt as 6 points (on the left) and bbox (on the right). Upperbound refers to the result of fine-tuning each class individually. The \textbf{bold} content is the highest value excluding the Upperbound.}\label{tab2}
\setlength{\tabcolsep}{1.3mm}{
\begin{tabular}{cccccc}
\hline
\multicolumn{1}{c|}{\multirow{2}{*}{Model}} & \multicolumn{4}{c|}{Specific category (Point/Bbox)}                                                                                       & \multirow{2}{*}{\begin{tabular}[c]{@{}c@{}}Weight mean\\ (Point/Bbox)\end{tabular}} \\ \cline{2-5}
\multicolumn{1}{c|}{}                       & \multicolumn{1}{c|}{Aorta}       & \multicolumn{1}{c|}{Stomach}     & \multicolumn{1}{c|}{Small bowel} & \multicolumn{1}{c|}{Esophagus}   &                                                                                     \\ \hline
\multicolumn{1}{c|}{Baseline}               & \multicolumn{1}{c|}{0.517/0.632} & \multicolumn{1}{c|}{0.442/0.500} & \multicolumn{1}{c|}{0.362/0.398} & \multicolumn{1}{c|}{0.348/0.464} & 0.447/0.523                                                                         \\ \hline
\multicolumn{1}{c|}{Upperbound}             & \multicolumn{1}{c|}{0.792/0.755} & \multicolumn{1}{c|}{0.717/0.687} & \multicolumn{1}{c|}{0.545/0.533} & \multicolumn{1}{c|}{0.593/0.566} & 0.687/0.660                                                                         \\ \hline
\multicolumn{6}{c}{Variations in $\tau$}                                                                                                                                                                                                                      \\ \hline
\multicolumn{1}{c|}{$\tau_{0.5}$}           & \multicolumn{1}{c|}{\textbf{0.632}/0.647} & \multicolumn{1}{c|}{\textbf{0.587}/\textbf{0.595}} & \multicolumn{1}{c|}{0.326/0.478} & \multicolumn{1}{c|}{0.391/\textbf{0.549}} & \textbf{0.522}/\textbf{0.589}                                                                         \\
\multicolumn{1}{c|}{$\tau_{0.7}$}           & \multicolumn{1}{c|}{0.597/0.638} & \multicolumn{1}{c|}{0.554/0.562} & \multicolumn{1}{c|}{\textbf{0.374}/0.429} & \multicolumn{1}{c|}{0.404/0.523} & 0.508/0.565                                                                         \\ \hline
\multicolumn{6}{c}{Weighted Approach}                                                                                                                                                                                                                         \\ \hline
\multicolumn{1}{c|}{Avg}                    & \multicolumn{1}{c|}{0.590/\textbf{0.661}} & \multicolumn{1}{c|}{0.503/0.590} & \multicolumn{1}{c|}{0.219/\textbf{0.484}} & \multicolumn{1}{c|}{\textbf{0.405}/0.538} & 0.480/0.588                                                                         \\
\multicolumn{1}{c|}{$Aft_{weight}$}         & \multicolumn{1}{c|}{0.027/0.521} & \multicolumn{1}{c|}{0.073/0.532} & \multicolumn{1}{c|}{0.116/0.443} & \multicolumn{1}{c|}{0.004/0.325} & 0.040/0.458                                                                         \\ \hline

\end{tabular}
}
\end{table}

\section{Conclusion}
This paper introduces a plug-and-play MoE framework based on SAM-Med3D, which seamlessly integrates task-specific finetuned models with the foundational model, creating a unified model at minimal additional training expense for an extra gating network. Then, a following selection strategy is adopted to enable the unified model to achieve comparable performance of the original models in their respective tasks without updating any parameters. Extensive experiments on 15 specific classes and the new SPPIN task demonstrate the effectiveness of SAM-Med3D-MoE. In future work, we will focus on two potential problems: (1) We will verify the effectiveness of more foundation models for medical image segmentation; (2) The hype-parameter $\tau$ in the mask selector should be dynamically adapted to any scenarios.

\begin{credits}
\subsubsection{\ackname} This research was supported by Shanghai Artificial Intelligence Laboratory.
\subsubsection{\discintname}
The authors have no competing interests to declare that are
relevant to the content of this article.
\end{credits}
\newpage
\bibliographystyle{splncs04}
\bibliography{Paper-3979}
\end{document}